\documentclass[letterpaper, 10 pt, conference]{ieeeconf}  
\usepackage[english]{babel}
\IEEEoverridecommandlockouts                              
\usepackage[noadjust]{cite}
\usepackage{algorithm,algorithmic}
\usepackage{amssymb}
\usepackage{amsmath}
\usepackage{amsfonts}
\RequirePackage{tcolorbox}
\newcommand{\Co}[1]{\text{\tiny\color{blue}\texttt{#1}}}
\newcommand{\Coor}[1]{\text{\color{blue}\texttt{#1}}}
\newcommand{\Exp}{\operatorname{Exp}}
\newcommand{\Log}{\operatorname{Log}}
\newcommand{\R}{\mathbb{R}}
\newcommand{\se}{\mathfrak{se}}
\newcommand{\so}{\mathfrak{so}}
\newcommand{\SO}{\operatorname{SO}}
\newcommand{\SE}{\operatorname{SE}}
\DeclareMathOperator*{\argmin}{arg\,min}
\newcommand{\ve}[1]{\boldsymbol{#1}}

\newcommand{\indizes}[2]{\,^\Co{#1}_\Co{#2}}
\newcommand{\indec}[1]{\,^\Co{#1}}
\usepackage{graphicx}
\usepackage{subcaption}
\usepackage[font={bf}]{caption}

\usepackage{hyperref}

\title{\LARGE \bf
Constrained Visual-Inertial Localization With Application And Benchmark in Laparoscopic Surgery
}
\author{Regine Hartwig$^{1,3}$, Daniel Ostler$^{1}$, Jean-Claude Rosenthal$^{2}$\\Hubertus Feu{\ss}ner$^{1}$, Dirk Wilhelm$^{1}$, Dirk Wollherr$^{3}$
\thanks{$^{1}$Authors are with Research Group MITI, Surgical Department, Technical University of Munich (TUM), Germany}%
\thanks{$^{2}$Authors are with Fraunhofer HHI, Berlin, Germany}
\thanks{$^{3}$Authors are with Chair of Automatic Control Engineering, Department of Electrical and Computer Engineering, Technical University of Munich (TUM), Germany {\tt\small regine.hartwig@tum.de}}
}
\begin{document}

\maketitle
\thispagestyle{empty}
\pagestyle{empty}
\begin{abstract}
We propose a novel method to tackle the visual-inertial localization problem for constrained camera movements. 
We use residuals from the different modalities to jointly optimize a global cost function. 
The residuals emerge from IMU measurements, stereoscopic feature points, and constraints on possible solutions in $\SE(3)$.
In settings where dynamic disturbances are frequent, the residuals reduce the complexity of the problem and make localization feasible. 
We verify the advantages of our method in a suitable medical use case and produce a dataset capturing a minimally invasive surgery in the abdomen.
Our novel clinical dataset MITI is comparable to state-of-the-art evaluation datasets, contains calibration and synchronization and is available \mbox{at \cite{hartwig2021dataset}}.
\end{abstract}

\section{INTRODUCTION}
Many researchers have studied simultaneous localization and mapping (SLAM) in the computer vision, and robotics community \cite{cadena2016past}, inspiring many innovative products.
Applications such as virtual reality, augmented reality \cite{klein2007parallel}, unmanned vehicles, and autonomous robots \cite{janai2020computer} rely heavily on stable pose estimates based on SLAM systems.

Challenging situations include a dynamic environment which prevents the SLAM algorithms from being applicable in many use-cases.
The works of \cite{forster2016svo, kerl13icra, leutenegger2015keyframe}  show that it is possible to add residuals specially designed for a use-case and jointly optimize a cost function.
Motion priors, reprojection from a calibrated multi-camera setup, or additional measurements reduce the degrees of freedom on the solution space.

MIS has proven to have many medical advantages for the patient.
Stable virtual/ augmented reality and assisting roboter arms could lead to enhanced hand-eye coordination in MIS and subsequently to a whole new way of conducting surgeries. Furthermore, SLAM is a game-changer for applications such as mapping detected metastasizing carcinoma for cancer staging, safety routines (e.g., collision avoidance), autonomous systems, image-guided surgery, input generation for artificial intelligence, and anatomic scene graphs.
The huge disadvantages (magnetic distortion, line-of-sight, room consumption) disqualify available tracking equipment for most interventions.
Data from this domain was acquired, and implementation shows the applicability of the proposed constrained visual-inertial SLAM method in this challenging dynamic surrounding.

The paper is structured as follows: Section \ref{sec_method} presents the visual-inertial odometry algorithm with movement constraints. Section \ref{sec_dataset} discusses our novel dataset \cite{hartwig2021dataset} and provides useful information for users.
In Section \ref{sec_eval} we will discuss the implementation of the proposed algorithm, discuss properties of the calibrated data from our dataset, and evaluate the performance by experiments.

\section{PROBLEM STATEMENT AND CONTRIBUTION}
\paragraph{Constrained Visual-Inertial Localization}

\cite{klein2007parallel}, and further developments have become popular state-of-the-art methods \cite{mur2017orb}. These algorithms focus on feature point matching based on binary descriptors, even for small viewpoint changes, where descriptor matching is made real-time applicable by speeding up the process with priorly trained or online learned \cite{7139959} bag-of-words vocabularies \cite{galvez2012bags,nister2006scalable}.

However, we focus on domains where feature point descriptor matching and tracking perform worse due to challenging image properties \cite{6820756}. Sparse or semi-dense optical flow \cite{fu2020fastorb,forster2016svo,engel2017direct} are alternatives that perform well for
high video framerate and can speed up trajectory generation for the \textit{tracking} part of the SLAM system.

Additional information from inertial measurement units (IMU) \cite{von2018direct,qin2018vins,qin2019general} improves visual odometry methods in difficult situations, where tracking is lost, and does not need an external measurement equipment, i.e., avoiding line-of-sight problems.


In contrast to all previously mentioned strategies, we additionally exploit naturally given motion constraints \cite{6907267}, which is also famous in the medical domain as remote center-of-motion (RCM) \cite{vasconcelos2019rcm}\cite{lin2016video}, reducing the solution space to tackle the more challenging image domains, which contain large movements in the image, occlusions, fog, deformable surfaces or difficult illuminations conditions \cite{lin2016video}.

Subsequently, we outline our approach in more detail:
We formulate residuals for IMU, stereo camera, and movement constraints and show that the solution space reduces by pivoting the camera around a point from 6 to 4 degrees of freedom. Afterward, we propose to use optical flow in combination with gyroscope readings as an odometry alternative to descriptor-based feature tracking for dynamic image data and high video frequency. Finally, for the refinement of this odometry, we formulate the global cost function and optimization strategies.
Our method's main advantage and contribution is the ability to relocalize the camera in unseen environments by minimizing the global cost function, which incorporates all residual terms emerging from sensors and constraints.
The gravity and magnetic field provide the necessary absolute references. In addition, the Lucas-Kanade tracking of the stereo reconstructed locally rigid parts in the environment provides information about the distance from the pivot point.

\paragraph{Dataset}
Publicly available datasets play a vital role in the evolution of
new algorithms, and technologies \cite{baker2011database,geiger2012we,sturm2012benchmark,allan2021stereo}. 
Researchers focusing on the minimally invasive application have already created endoscopic datasets \cite{981231,lin2016video,allan2021stereo}, which are suitable for investigating computer tomography registration, tissue deformation/ reconstruction \cite{8410942, mountney2010three, maier2013optical}, and instrument segmentation. 
Since they are not complete or are not capturing a handheld intervention,
those are unsuitable for testing SLAM applications. 
In this paper, we introduce a novel dataset suitable for testing SLAM in the MIS domain. To the best of our knowledge, our dataset is the first publicly available surgical dataset comparable to \cite{baker2011database,geiger2012we,sturm2012benchmark}.
Our MITI Dataset \cite{hartwig2021dataset}, available under the Creative Commons Attribution license (CC-BY 4.0), provides all necessary data by a complete recording of a handheld surgical intervention at Research Hospital Rechts der Isar of TUM. It contains multimodal sensor information from IMU, stereoscopic video, and infrared (IR) tracking as ground truth for evaluation.
Furthermore, calibration for the stereoscope, accelerometer, magnetometer, the
rigid transformations in the sensor setup, and time-offsets are available. 
We wisely chose a suitable intervention, namely diagnostic laparoscopy, that contains very few cutting and tissue deformation and shows a full scan of the abdomen with a handheld camera such that it is ideal for testing SLAM algorithms.
It incorporates 667 seconds with images (1080p60 RGB) for each camera, readings of two IMU sensors at 220Hz, and IR transformations to two different targets at 20Hz.

Intending to promote the progress of visual-inertial algorithms designed for MIS application, we hope that our clinical training dataset helps and enables researchers to enhance algorithms.

\section{NOTATION}
The following notions are adapted from \cite{barfoot2017state,sola2018micro,sommer2020efficient} using the rotation groups $\SO(3)$ and the rigid motion group $\SE(3)$, which fulfill the properties of a Lie group. We use the notation  $\exp: \mathfrak{m} \to \mathcal{M}$ to denote the mapping from tangent space, i.e. Lie algebra $\mathfrak{m}$ to Lie group $\mathcal{M}$.
The Lie algebra $\so(3){=}\operatorname{skew}(3)$ of $\SO(3)$ has $d{=}3$ degrees of freedom and $\se(3)$, tangent space of $\SE(3)$, has $d{=}6$ degrees of freedom. They are isometric isomorph to $\R^d$, and we use the function $\Exp$ as the composition of the hat operator and the matrix exponential $\exp$. The function can explicitly be given in closed form by the Rodriguez formula.
The inverse mapping is denoted by $\Log$, which is the mapping from the Lie-Group to $\R^d$.
We further use the $\ominus:\mathcal{M}\times \mathcal{M} \to \R^d$  as
\begin{equation}
\mathbf{T}_1 \ominus \mathbf{T}_2:=\Log\left( \mathbf{T}_1^{-1} \mathbf{T}_2 \right),
\end{equation}
which computes the difference between two transformation matrices mapped to the tangent spaces' isomorph space $\R^d$.
\section{VISUAL-INERTIAL SENSOR FUSION}\label{sec_method}
Our method splits into two parts: Firstly, estimating the pose starting from previous keyframes by movement constraints, gyroscope readings, and Lucas-Kanade tracks of stereoscopic matches. Secondly, the algorithm conducts a windowed optimization over gathered keyframes, changing their location by jointly minimizing the residuals from all modalities, eliminating any drift that might accumulate.

\subsection{Residuals}\label{sec_method_residuals}
This subsection describes the residuals gathered from different modalities. The residuals are functions of the localization pose $\indizes{C}{T}\mathbf{T} \in \SE(3)$ with the property that they are zero at the actual position and orientation.

\paragraph{Motion Constraints}
The transformation $\indizes{C}{T}\mathbf{T} \in \SE(3)$ transforms a vector in homogeneous coordinates from \Coor{T} to \Coor{C} by multiplication from the right. Since the motion of \Coor{C} is constrained to move around pivot point \Coor{T} but can also move forward and backward in the x-direction, we can reduce the 6 degrees of freedom to 4 degrees of freedom 
\begin{equation}
\indizes{C}{T}\mathbf{T} =
\begin{bmatrix}
\indizes{C}{T}\mathbf{R} & \indizes{C}{T}\ve{t}\\
\ve{0}^T & 1
\end{bmatrix} = 
\begin{bmatrix}
I & \indizes{C}{T}\ve{t}\\
\ve{0}^T & 1
\end{bmatrix}
\begin{bmatrix}
\indizes{C}{T}\mathbf{R} &\ve{0}\\
\ve{0}^T & 1
\end{bmatrix},
\end{equation}
with $\indizes{C}{T}\ve{t} = t_x  \ve{e}_x$. This means a point in \Coor{T} is first rotated in the pivot point and then translated along the x-axis.
The corresponding residual that enforces the result to have only 4 degrees of freedom is
\begin{equation}\label{eq_respivot}
\ve{r}_{pivot} =  t_{y} \ve{e}_y + t_{z}\ve{e}_z.
\end{equation}
This adds a penalty to the minimization problem if the solution is not in the reduced space.
\paragraph{Accelerometer}
The accelerometer readings constrain the current camera pose by one corresponding bearing vector in world coordinates.
We assume constant velocity as also proposed by \cite{madgwick2010efficient}.
Rotating the gravity $\indec{T}\ve{g}$ bearing vector from \Coor{T} to \Coor{C} and comparing it with the   measurement $\indec{C}\ve{g}$ gives the residual
\begin{equation}\label{eq_resa}
\ve{r}_a =\indec{C}\ve{g} - \indizes{C}{T}\mathbf{R}  \indec{T}\ve{g}.
\end{equation}
\paragraph{Magnetometer}
The magnetometer provides similar to the accelerometer a correspondence in world coordinates
\begin{equation}\label{eq_resm}
\ve{r}_m = \indec{C}\ve{m}-\indizes{C}{T}\mathbf{R}  \indec{T}\ve{m},
\end{equation}
with the magnetic field $\indec{T}\ve{m}$ in \Coor{T} and measurement $\indec{C}\ve{m}$ \mbox{in \Coor{C}}.
\paragraph{Global Mappoints}
The distance between a mappoint $^\Co{T}\ve{x}$ in global coordinates and a matched 2D position $\ve{u} $ in image plane of current camera coordinates is
\begin{equation}\label{eq_res2D}
\ve{r}_{2D} = \Pi(\indizes{C}{T}\mathbf{T}  \, \indec{T}\ve{x}) -
\ve{u}
\end{equation}
with $\Pi(\cdot)$ being the projection into distorted image plane and $\ve{u}{:=}\ve{u}_d$ the distorted image coordinates.
In (\ref{eq_res2D}), the map points $^\Co{T}\ve{x}$ are connected rigidly to the world coordinate system \Coor{T} at initialization by using the current camera pose estimate. We need to rewrite (\ref{eq_res2D}), attaching the mappoints to the camera coordinate system. Otherwise, since we optimize only the poses $\indizes{C}{T}\mathbf{T}$, the minimization of other residuals leads to the increase of (\ref{eq_res2D}). 
For this sake, we store the landmark positions relative to the camera coordinate system \Coor{C}, in which the stereo matcher initialized them. 
This adds another computation step and one more parameter $ \indizes{T}{C}\mathbf{T}_{t'}$  at detection time $t'$ to the residual in (\ref{eq_res2D}), namely
\begin{equation}
\indec{T}\ve{x} = \indizes{T}{C}\mathbf{T}_{t'} \indec{C}\ve{x}_{t'}.
\end{equation}
Thus we have
\begin{equation}\label{eq_res2D_new}
\ve{r}_{2D} = \Pi(\indizes{C}{T}\mathbf{T}  \indizes{T}{C}\mathbf{T}_{t'} \indec{C}\ve{x}_{t'}) -
\ve{u},
\end{equation}
which connects the observation frame with the initialization frame of a landmark.

\paragraph{Gyroscope}
The gyroscope measurements in tangent space of  $\ve{\omega}_{t'} \in \R^3 \equiv \so(3)$ are integrated by gyroscope sample time $\Delta t$ and the non-abelsh product
\begin{equation}\label{eq_g}
\indec{C}\Delta \mathbf{R}_{t_1,t_2} := \prod_{t' \in \{t_1,...,t_2-\Delta t\}} \Exp(\indec{C}\ve{\omega}_{t'} \Delta t).
\end{equation}
This gives the update rule
\begin{equation}
\indizes{T}{C}\mathbf{R}_{t_2} = \indizes{T}{C}\mathbf{R}_{t_1} \, \indec{C}\Delta \mathbf{R}_{t_1,t_2}
\end{equation}
and the residual term
\begin{equation}\label{eq_resg}
\ve{r}_g = \indizes{T}{C}\mathbf{R}_{t_2} \ominus \indizes{T}{C}\mathbf{R}_{t_1} \, \indec{C}\Delta \mathbf{R}_{t_1,t_2}.
\end{equation}
which constraints consecutive transformations to have a certain deviation in orientation, namely the integrated measurements of the angular velocity.
\subsection{Visual-Inertial Odometry Tracking}\label{sec_method_tracking}
This subsection describes our approach for the odometry tracking thread. It updates from the pose of the previous keyframe, generating good initializations for the optimization using gyroscope data and Lucas Kanade tracks of stereo correspondences.
Algorithm \ref{alg1} sketches the procedure, and in the following, we will explain line by line the computation steps the algorithm conducts.\par 

After entering the loop, it first updates the pose based on the gyroscope measurements in Line \ref{alg1_update_gyro}. 
Although (\ref{eq_resg}) usually only provides information about the orientation, a position update is  possible due to the camera movement having only 4 degrees of freedom
\begin{equation}\label{update_imu}
\indizes{C}{T}\mathbf{T}_{t_{2} }= 
\begin{bmatrix}
I & \indizes{C}{T}\ve{t}_{t_1}\\[4pt]
\ve{0}^T & 1
\end{bmatrix}
\begin{bmatrix}
\indec{C}\Delta \mathbf{R}_{t_1,t_2}^T \,\indizes{C}{T}\mathbf{R}_{t_1} &\ve{0}\\[4pt]
\ve{0}^T & 1
\end{bmatrix}
\end{equation}
with $\indizes{C}{T}\ve{t}_t = t_{x,t} \ve{e}_x$. The new pose is on a sphere with radius $t_{x,t}$, around the trocar entry point, leaving only the radius unknown but estimating all other degrees of freedom.\par
During the first iteration, the Lines \ref{alg1_update_points_lk} to \ref{alg1_update_poses_lk} have no effect.
The first frame is a keyframe, and the algorithm computes stereo correspondences by descriptor matching using ORB descriptors \cite{rublee2011orb} in Line \ref{alg1_stereo}.
ORB descriptors are faster to compute and compare than SIFT or SURF, whose benefits, e.g., scale and rotation invariance, are less essential for stereo-matching.
For faster matching, it uses the projections provided by the camera calibration parameters as initialization for finding matches in the second image \cite{mur2017orb}. The epipolar error threshold \cite{forsyth2011computer} is adjusted to filter outliers
\begin{equation}
\indec{C0}\ve{x}^T \mathbf{E} \indec{C1}\ve{x} > d_{th},
\end{equation}
with epipolar matrix $\mathbf{E}= \hat{\ve{t}}\mathbf{R}$
and $\hat{\cdot}:\R^3 \to\text{skew}(3)$, defining the transformation between left  \Coor{C0} and right \Coor{C1} lens.
The matched points initialize new landmarks in Line \ref{alg_1_insert_lm}.\par
Afterwards, in Line \ref{alg1_update_points_lk} for every incoming frame it updates the landmark projections $\ve{u} \in \R^2$ in the 2D image plane, producing residuals of the form (\ref{eq_res2D}). For that, it uses the variational Lucas-Kanade-Method \cite{bruhn2005lucas} taking two consecutive frames for every camera. This variational method performs well (small baseline, photometric consistency) if video frequency is high. 
For large displacements, the drift becomes large, and for fast movements, we lose track. Nevertheless, optical flow is suitable for local tracking.
\par
With initialization computed in Line \ref{alg1_update_gyro} it determines the camera pose  $\indizes{C}{T}\mathbf{T}_t$ based on the 3D-2D-correspondences in Line \ref{alg1_update_poses_lk} by a perspective-n-point algorithm \cite{lepetit2009epnp} inside a random sample consensus (RANSAC) \cite{fischler1981random} scheme with refinement provided by OpenGV \cite{kneip2014opengv}. A sanity check on the rotation part using the IMU measurements and on the translation part using the constraints ensures locally rigid features.
\begin{algorithm}[] 
\caption{Visual-Inertial Odometry with Optimization}\label{alg1}
\begin{algorithmic}[1]
\renewcommand{\algorithmicrequire}{\textbf{Input:}}
\renewcommand{\algorithmicensure}{\textbf{Output:}}
\REQUIRE Initialization $\indizes{C}{T}\mathbf{T}_{t_{0}}$, Images $\{I_{0,t_i},I_{1,t_i}\}_{i \in I}$; \\IMU Data $\{\indec{C}\Delta \mathbf{R}_{t_i,t_{i+1}}, \indec{C}\ve{a}_{t_i}, \indec{C}\ve{m}_{t_i}\}_{i\in I}$
\ENSURE $\{\indizes{C}{T}\mathbf{T}_{t_i}\}_{i\in I}$
\WHILE {True}
\STATE Update $\indizes{C}{T}\mathbf{T}_{t_{i-1}}$ based on $\indec{C}\Delta \mathbf{R}_{t_{i-1}}$ and movement constraints\label{alg1_update_gyro}
\STATE Lukas-Kanade tracking of landmarks\label{alg1_update_points_lk}
\STATE Update $\indizes{C}{T}\mathbf{T}_{t_{i}}$ based on 2D-landmark projections\label{alg1_update_poses_lk}
\IF{Decision for keyframe}
  	\STATE Stereo matching \label{alg1_stereo}
  	\STATE Landmark insertion\label{alg_1_insert_lm}
\ENDIF
\STATE $i = i + 1$
\ENDWHILE
 \end{algorithmic} 
 \end{algorithm}

\subsection{Optimization}\label{sec_method_optimization}

The optimization presented in this subsection jointly minimizes all previously formulated residuals over a window of keyframes. 
As shown later the camera can globally relocalize itself by solving the proposed optimization problem solely by small tracks of locally rigid feature points and IMU measurements.
In order to use all the information from sensor readings and images we build a cost function by summing up the normalized, weighted and robustified residuals that have been collected, i.e. (\ref{eq_respivot}, \ref{eq_resa}, \ref{eq_resm}, \ref{eq_res2D_new}, \ref{eq_resg}) which are available at each timestep.
We denote them as residuals 
$\ve{r}_k$, $k\in K$, where $k$ indicates time and residual type.

The overall optimization problem is
\begin{equation}\label{eq_min}
\argmin_{\indizes{C}{T}\mathbf{T}_{n:m}} \quad \sum_{k}
\rho_k \left( \left\| \alpha_k\ve{r}_k\left(\indizes{C}{T}\mathbf{T}_{n:m} \right) \right\| \right){:=}\sum_{k} f_k(\indizes{C}{T}\mathbf{T}_{n:m}),
\end{equation}
optimizing over the window of keyframe localizations $\indizes{C}{T}\mathbf{T}_{n:m} := \left( \indizes{C}{T}\mathbf{T}_{t_n},...,\indizes{C}{T}\mathbf{T}_{t_m} \right)$.
In above formulation, $\alpha_k$ stands for the scaling factor, and $\rho$ for the robustification function.
Each residual gets an individual scaling factor $\alpha_k$. 
The evaluation part of this paper gives a more detailed overview of the chosen weight factors, optimization window size, and trigger of the optimization problem.\par
For robustification of residuals that are prone to outliers, we choose the Huber loss 
\begin{equation}
\rho_k(x)  = \begin{cases}
 \frac{1}{2}{x^2}                   & \text{for } |x| \le \delta, \\
 \delta ( | x | - \frac{1}{2}\delta), & \text{otherwise.}
\end{cases}
\end{equation}
The function input $x$ in our application is the normalized and weighted residual $\| \alpha_k r_k \|$.
For residuals with a dense error distribution and no outliers, $\rho_k$ is simply the identity. \par
Each of the residuals $\ve{r}_k\left(\indizes{C}{T}\mathbf{T}_{n:m}\right)$ depends only on a maximum of two parameters $\indizes{C}{T}\mathbf{T} \in \SE(3)$, and further, many parameters do not have a residual in common.
The residual terms give the optimization problem the typical graph structure of SLAM problems.
This sparse dependence on parameters makes the Hesse-matrix sparse, and the optimization problem suitable for a sparse Schur solver \cite{grisetti2011g2o}.

\section{DATASET}\label{sec_dataset}
\begin{figure}[b!]
\begin{subfigure}[c]{0.27\textwidth}
\centerline{\includegraphics[height = 0.08\textheight]{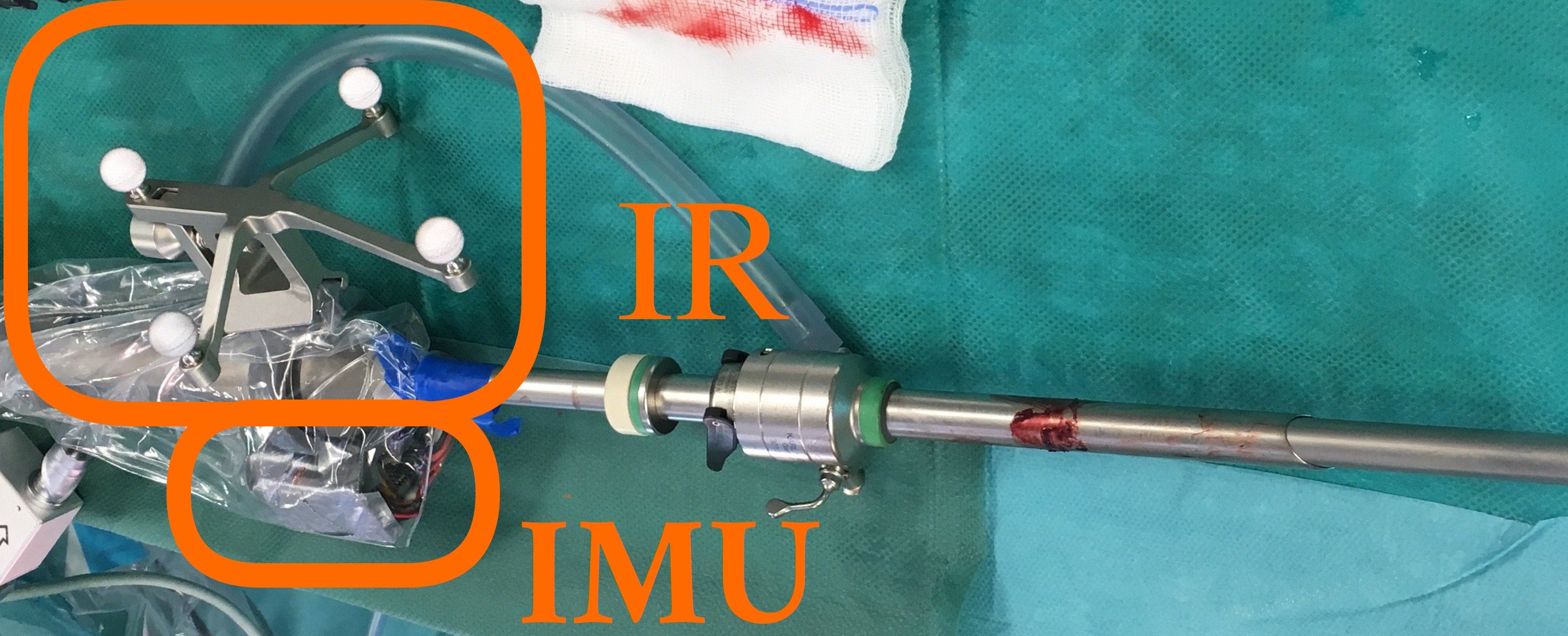}}
\subcaption{}\label{fig_1_subfigure_1}
\end{subfigure}
\begin{subfigure}[c]{0.18\textwidth}
\centerline{\includegraphics[height = 0.08\textheight]{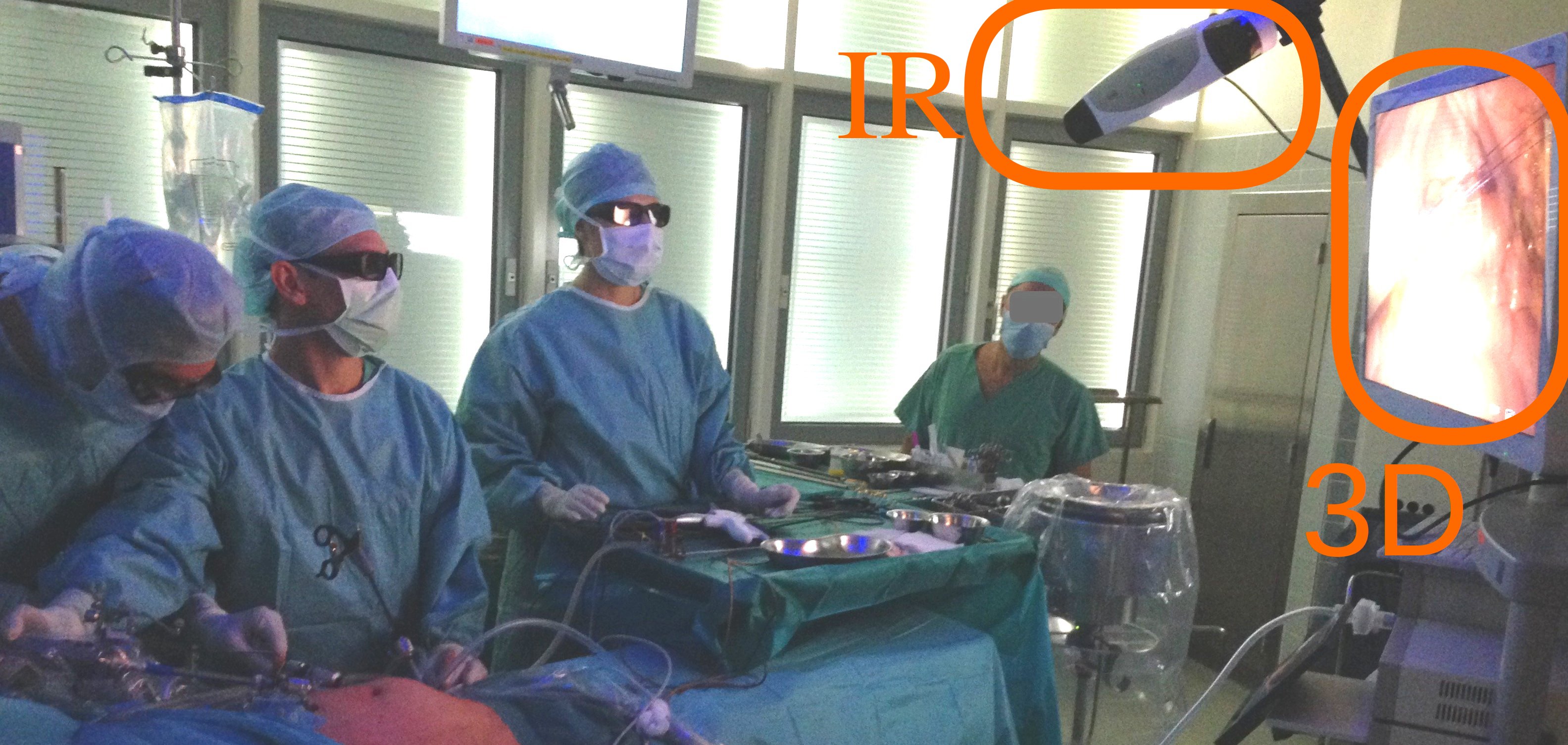}}
\subcaption{}\label{fig_1_subfigure_2}
\end{subfigure}
\caption{
Data Acquisition (a) IMU 
sensorboards
 and IR passive targets attached to stereoscopic laparocopic camera (b) IR sensor and 3D monitor in the operating room}
\label{fig1}
\end{figure}

In this section, we introduce our dataset and calibration procedures. 
In  Fig. \ref{fig1} we show the sensor setup during the data acquisition process.
The sensor data consists firstly of a stereoscopic 
video (Tipcam1, Karl Storz) from the view inside the abdomen.
Secondly, two 9DOF IMU (MetamotionR, Metawear) sensor boards attached to the laparoscope handle sending:
angular velocity, acceleration, and magnetic field strength. 
Thirdly, an IR tracking camera (Polaris Vega, NDI) 
detects two targets, each consisting of 3-4 reflecting passive spheres.
The anticipated precision of the IR tracking is 0.25mm \cite{fattori2021technical}, making it suitable as ground truth to evaluate the SLAM algorithm's performance.
We calibrated the camera and the IMU, determining sensor-specific parameters and calculated time offsets using the state-of-the-art methods defined in subsequent subsections.
Parameters can be found in the path \texttt{calibration/*.csv} of the dataset \cite{hartwig2021dataset}.
\setlength{\textfloatsep}{0pt}
\subsection{Camera Calibration}
The pinhole projection \cite{zhang2000flexible} $\Pi_0: \R^3 \to\R^2 $
$\ve{x} \mapsto \ve{u}_u$
 from a point $\ve{x} \in \R^3$ in coordinate system of left \Coor{C0} or right \Coor{C1} lens to the undistorted image point $\ve{u}_u$ in the image plane is parameterized by focal lengths $f_x,f_y$ and optical center $c_x,c_y$  both measured in $\mathrm{pixel}$.


Radial image distortion is compensated by a low order polynomial model 
$d: \R^2 \to \R^2 $, $\ve{u}_u \mapsto \ve{u}_d$.
Additionally, the extrinsic parameters are calibrated, which determine 
the rigid body transformation $\indizes{C0}{C1}T \in \SE(3)$.
The function $\Pi(\cdot)$ in residual (\ref{eq_res2D}) is a composition of rotation and translation from \Coor{C} to \Coor{C0} or \Coor{C1} and projection 
to distorted 2D image coordinates by composition of
$\Pi_0$ and $d$.
(\ref{calib_cam_radial}).
\subsection{IMU Calibration}
\begin{figure}[tp]
\begin{subfigure}[c]{0.19\textwidth}
\centerline{\includegraphics[height = 0.08\textheight]{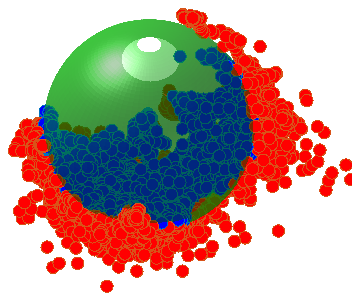}}
\subcaption{}\label{fig_accel_readings} 
\end{subfigure}
\begin{subfigure}[c]{0.19\textwidth}
\centerline{\includegraphics[height = 0.08\textheight]{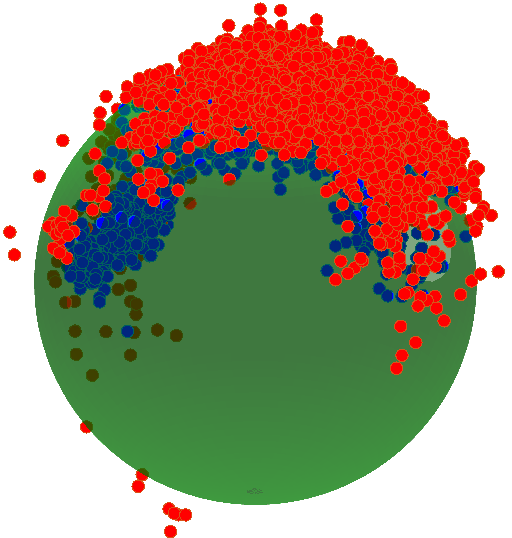}}
\subcaption{}\label{fig_mag_readings}
\end{subfigure}
\caption{Calibrated IMU measurements, i.e., red in front, blue behind sphere (a) $\indec{C}\ve{a}-\indec{C}\ve{g}$ in $9.81 \frac{m}{s^2}$ and unit sphere (b) $\indec{C}\ve{m}$ in $\mu T$ and sphere with radius $48.6$}\label{fig_mag_accel_readings}
\end{figure}

Using a standard approach \cite{madgwick2010efficient,renaudin2010complete}, the accelerometer measurements $\tilde{\ve{a}}(t)$ are modeled  as 
\begin{equation}
\tilde{\ve{a}}(t) =  (\ve{a}(t) - \ve{g}(t) ) d_a^{-1} + \ve{n}_a(t) + \ve{b}_a
\end{equation}
with acceleration $\ve{a}(t)\in \R^3$, gravity $\ve{g}(t)\in \R^3$, errors $\ve{b}_a \in \R^3$, $d_a \in \R$ and $\ve{n}_a(t) \sim  \mathcal{N}(0,\sigma_a ^2\mathbf{I})$. Calibration assumes constant velocity and gravity norm $g = \| \ve{g}(t)\| = 9.81 \frac{m}{s^2}$.
Furthermore, we model the error of magnetometer readings $\tilde{\ve{m}}(t)$ \cite{madgwick2010efficient,renaudin2010complete} by
\begin{equation}
\tilde{\ve{m}}(t) =   \ve{m}(t)d_m^{-1} + \ve{n}_m(t) + \ve{b}_m,
\end{equation}
with magnetic field strength $\ve{m}(t)\in \R^3$, errors $\ve{b}_m \in \R^3$, $d_m \in \R$ and $\ve{n}_m(t) \sim  \mathcal{N}(0,\sigma_m ^2\mathbf{I})$. 
The corrected measurements lie on a sphere with radius equal to the strength of the earths magnetic field $m = \|\ve{m}\| = 48.6 \mu T$ in Munich, Germany. 
This information is used in the calibration procedure.
Fig. \ref{fig_mag_accel_readings} shows calibrated sensor readings.
\subsection{Time Offset Calibration}
The sensor data has been synchronized during the acquisition process by configuring a network time protocol server/client at the data receiving devices. 
\begin{figure}[t]
\centerline{\includegraphics[width = 0.4 \textwidth]{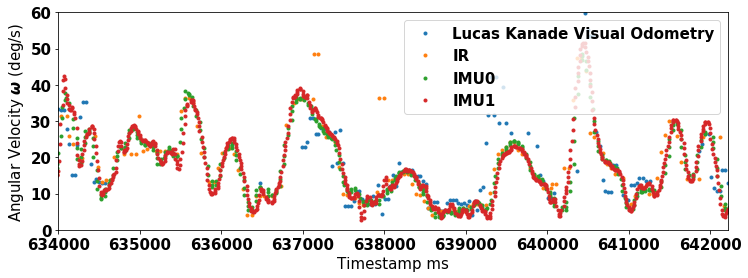}}
\caption{Time-calibrated angular velocity part $\|\indec{C}\ve{\omega}^{(k)}_{t_i}\|$}\label{fig3}
\end{figure}For synchronization of the video the displayed timestamp has been captured at the start and end of the intervention. The sensor interfaces for IR/IMU are ethernet/bluetooth respectively, which causes a remaining timeoffset in the collected data. This offset is minimized by evaluating
the tangent $ \indec{C}\ve{\xi}_{t_i} \in \R^6 \cong \se(3)$ for each sensor at coordinate system \Coor{C} at time $t_i$
We then calibrated the timeoffset $dt$ by comparing the velocities acquired from different sensors $$j,k \in \{\text{"Visual Odometry", "IR", "IMU0", "IMU1"}\}.$$ 
The sum of differences at different timeoffsets is
\begin{equation}
\min_{dt}\sum_i\left\|\indec{C}\ve{\xi}^{(j)}_{t_i+dt} - \indec{C}\ve{\xi}^{(k)}_{t_i}    \right\|.
\end{equation}
For that, we determine the camera-based angular velocity using the visual odometry algorithm based on Lucas Kanade feature tracks without IMU and IR. The computed timeoffsets are 100ms/166ms for the IR/IMU readings.
Time synchronization is vital for short-term odometry, unlike windowed optimization, where keyframes are more distant.
\section{EVALUATION AND EXPERIMENTS}\label{sec_eval}
We conduct experiments on an implementation of the approach described in Section \ref{sec_method}.
In Subsection \ref{sec_eval_weight} we examine the influence of weight distribution on residuals, and in Subsection \ref{sec_eval_robust} we discuss the advantages of utilizing robust loss functions.
Subsection \ref{sec_eval_vo} and \ref{sec_eval_opti} are devoted to the evaluation of our method with our MITI dataset. In doing that, we measure the deviation of the algorithms output to ground-truth poses obtained by IR tracking.

We used several libraries: OpenCV for keypoint detection, feature tracking, and reprojection; OpenGV \cite{kneip2014opengv} for triangulation, RANSAC, and nonlinear refinement; as well as Ceres \cite{agarwal2012ceres} and Sophus \cite{sophus-library} for solving the optimization problem.
For the experiment, we took the image sequence starting from timestamp 637273ms, which contains a fast movement from left to right with dynamic occlusions and enables us to demonstrate the algorithm's performance in a scenario where standard SLAM approaches lose tracking and can not relocalize in the unseen environment.
\subsection{Evaluation of Visual Odometry Tracking}\label{sec_eval_vo}
\begin{figure}[t!]
\centering
\begin{subfigure}[c]{0.45 \textwidth}
\centerline{\includegraphics[width = 1\textwidth]{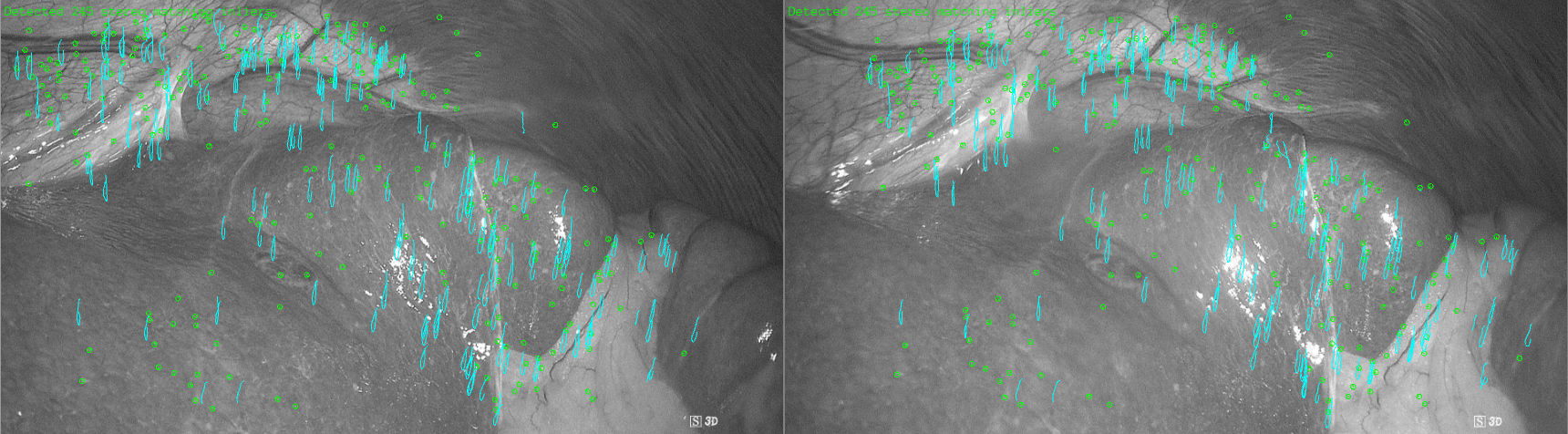}}
\subcaption{}\label{fig3Dview_subfigure_1}
\end{subfigure}
\begin{subfigure}[c]{0.23 \textwidth}
\centerline{\includegraphics[height = 0.08\textheight]{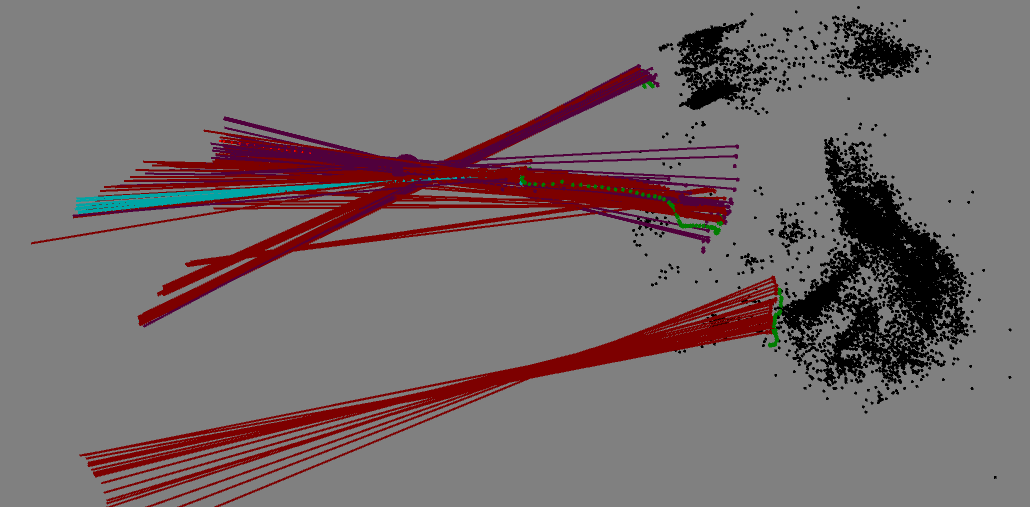}}
\subcaption{}\label{fig3Dview_subfigure_2}
\end{subfigure}
\begin{subfigure}[c]{0.22 \textwidth}
\centerline{\includegraphics[height = 0.08\textheight]{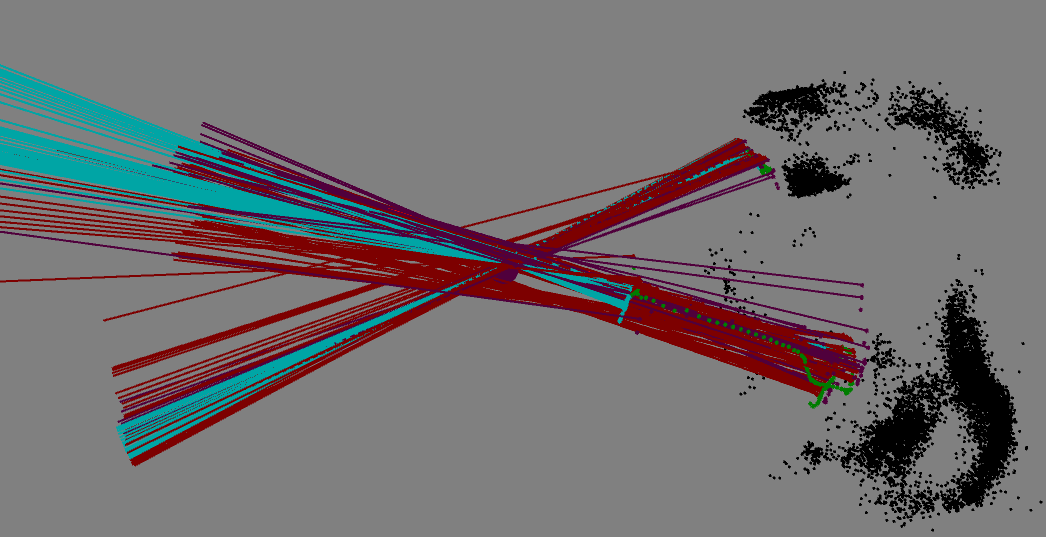}}
\subcaption{}\label{fig3Dview_subfigure_3}
\end{subfigure}
\caption{Debug environment (a) Stereo view with ORB-descriptor-based stereomatches in green, Lucas-Kanade-tracks in teal, (b) and (c) Camera poses connected to lines, that should go through pivot point and 3D map points of left and right diaphragm, (b) Camera-only based tracking, (c) Incorporating IMU sensordata and movement constraint}\label{fig3Dview}
\end{figure}

In Fig. \ref{fig3Dview} one can inspect the debug view of the output and Fig. \ref{fig3Dview_subfigure_2} and Fig. \ref{fig3Dview_subfigure_3} show huge deviations for different settings in the algorithm.
\begin{figure}[th!]
\centering
\begin{subfigure}[c]{0.40 \textwidth}
\centerline{\includegraphics[width = 1\textwidth]{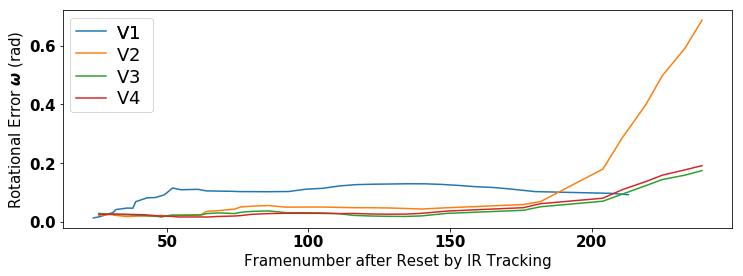}}
\subcaption{}\label{fig_4_subfigure_1}
\end{subfigure}
\begin{subfigure}[c]{0.40 \textwidth}
\centerline{\includegraphics[width = 1\textwidth]{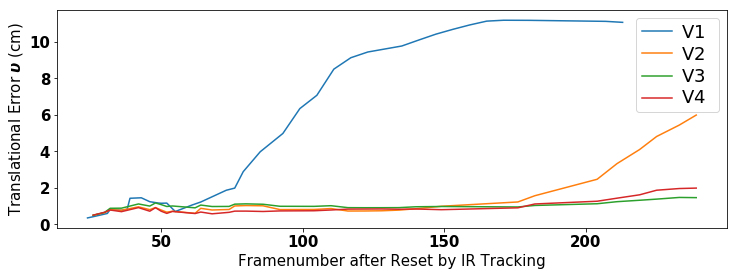}}
\subcaption{}\label{fig_4_subfigure_2}
\end{subfigure}
\caption{Algorithm \ref{alg1} without optimization (a) V1 relies on IMU data and can reduce rotational error in the long run, whereas V2 based on vision suffers from drift, (b) error of  V1 is unacceptable when movements do not stay on the sphere, error of pure visual odometry V2 also suffers from drift}\label{fig4}
\end{figure}
The isolated bundle of lines in Fig. \ref{fig3Dview_subfigure_2} shows: without movement constraints and IMU measurements, the optimization is unable to converge to the correct poses since it loses tracking of 2D features in between keyframes.
We computed the Log distance between IR tracked and constrained visual-inertial odometry (VO) estimated camera poses 
\begin{equation}
\ve{\xi}  = [\ve{\omega}, \ve{\upsilon}]^T = \indizes{C}{T}\mathbf{T}^{(\text{IR})} \ominus \indizes{C}{T}\mathbf{T}^{(\text{VO})} \in \R^6
\end{equation}

for  variants of the computation steps in Algorithm \ref{alg1}, that are listed in Table \ref{tab_fig4}.
V1 integrates IMU sensor-readings in Line \ref{alg1_update_gyro} into pose estimates as given in (\ref{update_imu}). Therefore, it cannot detect changes in the x-translation component, resulting in a large translational error as soon as the camera moves inside.
V2 only uses the camera-based pose estimation of Lines \ref{alg1_update_points_lk} and \ref{alg1_update_poses_lk} but without taking the initialization provided by the IMU in Line \ref{alg1_update_gyro}. The result of this variant can additionally be inspected in the debug view of Fig. \ref{fig3Dview_subfigure_2}.
V3 uses both camera-based and IMU-based tracking, which already leads to a significant decrease in both error residuals.
\begin{table}[h!]
\caption{Settings of Alg. \ref{alg1} for results shown in Fig. \ref{fig4}}\label{tab_fig4}
\begin{tabular}{l|l|l|l|l}
 Variant& Pivot point online & Line\ref{alg1_update_gyro}& Line \ref{alg1_update_points_lk} and \ref{alg1_update_poses_lk}  & Optimization \\
    \hline
V1 & $\times$ & \checkmark &  $\times$ &  $\times$\\
V2 & $\times$  & $\times$  & \checkmark &  $\times$\\
V3 &  $\times$ & \checkmark & \checkmark &  $\times$\\
V4 &  \checkmark & \checkmark & \checkmark &  $\times$\\
\end{tabular}
\end{table}
Finally, in V4, instead of considering the pivot point given by the priorly computed position from IR tracking, we initialize and iteratively refine the pose purely based on visual odometry.
The result is an essential step for the goal an eternal inside-out tracking.

\subsection{Error Analysis and Choice of Robustification Functions}\label{sec_eval_robust}
The residuals themselves are composed of errors in sensor readings and the deviation of the current pose estimate. Aiming to minimize only the part indicating a pose deviation, we examine errors of sensor readings and their propagation to the residual term to decide whether to use a Huber loss.
The Huber loss for the reprojection residual (\ref{eq_res2D_new}) is justified, since we identified an outlier characteristic in the three error sources: A drift error in the Lucas-Kanade tracked 2D locations $\indec{C}\ve{u}$ 
, that can be modeled as a random walk, after $n$ tracking steps, contributes with a Gaussian probability distribution of \mbox{$\ve{\epsilon}\sim\mathcal{N}(\ve{0},n\sigma^2 \mathbf{I})$}
linearly to the residual. The second error source is a badly reconstructed landmark position $\indec{C}\ve{x}_{t'}$,
 which occurs although the epipolar error threshold reduces the probability of outliers.
 The third error arises from a failing filtering of dynamic feature points. For residuals that penalize solutions not going through the trocar entry point (\ref{eq_respivot}) the use of the Huber loss is not justified, since there is no measurement error, and we aim to minimize large residuals. The error of IMU measurements (\ref{eq_resa}, \ref{eq_resm}, \ref{eq_g}) is approximately Gaussian, which leads us to not choosing the Huber loss.
\subsection{Weighting Residuals}\label{sec_eval_weight}
In this subsection we discuss the factor $\alpha_k$ that strongly influences the optimization (\ref{eq_min}).
Firstly, a task of the parameter is to adjust the residuals, making them independent on their measurement unit.
We ran Algorithm \ref{alg1} and built a statistic for every residual type, as shown in Tab \ref{tab_residuals_statistic}. 
\begin{table}[t]
\caption{Statistic of Residuals}\label{tab_residuals_statistic}
\begin{tabular}{l|l|l|l|l}
 &Eq. & Expectation $\operatorname{E}[ \ve{r}]$ & Variance $\operatorname{Var}[\ve{r}]$  & Unit\\
\hline
$\ve{r}_{2D}$ & (\ref{eq_res2D}) & 1.6263 & 9.3767 & $\mathrm{Pixel}$\\
$\ve{r}_{g}$ &(\ref{eq_g}) & 0.0003 & 0.0051 & $\mathrm{rad}$\\
$\ve{r}_{pivot}$ &(\ref{eq_respivot}) & 0.0511 & 0.4374 & $\mathrm{cm}$\\
$\ve{r}_{a}$ &(\ref{eq_resa})  & 0.0235 & 0.0405 & $9.81 \frac{\mathrm{m}}{\mathrm{s}^2}$\\
$\ve{r}_{m}$ &(\ref{eq_resm})  & -0.2730 & 4.2801 & $\mu \mathrm{T}$\\
\end{tabular}
\end{table}
\par
The expectation of the residuals is close to zero, which simplifies the normalization $\frac{\ve{r}-\operatorname{E}[\ve{r}_k]}{\sqrt{\operatorname{Var}[\ve{r}_k]}}$.
The scaling factor splits into $\alpha_k = \frac{1}{N_k}\frac{1}{\sqrt{\operatorname{Var}[\ve{r}_k]}} \beta_k$, with $\beta_k$ being an additional term to adjust the importance between residuals, as discussed in the next Subsection.
\begin{figure}[th]
\begin{minipage}[c]{0.39\textwidth}
\begin{subfigure}[l]{0.3\textheight}
\centerline{\includegraphics[width = 1\textwidth]{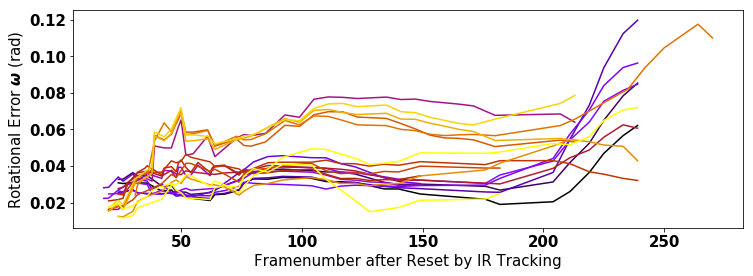}}
\subcaption{}\label{fig_residual_tradeoff_subfigure_1}
\end{subfigure}
\begin{subfigure}[l]{0.3 \textheight}
\centerline{\includegraphics[width = 1\textwidth]{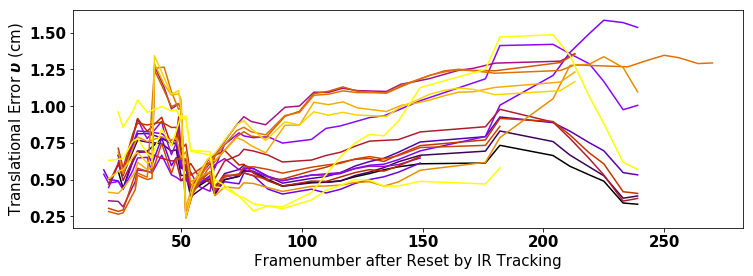}}
\subcaption{}\label{fig_residual_tradeoff_subfigure_2}
\end{subfigure}
\begin{subfigure}[l]{0.3 \textheight}
\centerline{\includegraphics[width = 1\textwidth]{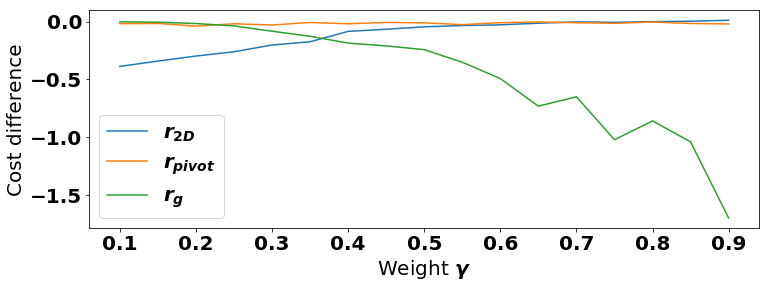}}
\subcaption{}\label{fig_residual_tradeoff_subfigure_3}
\end{subfigure}
\end{minipage}
\hfill
\begin{minipage}[t]{0.05\textwidth}
\centering
\includegraphics[width = 1\textwidth]{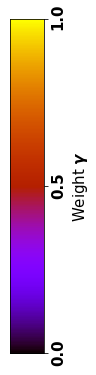}
\end{minipage}
\caption{Errors and cost reduction of V3 with optimization for different values of tradeoff value $\gamma$}\label{fig_residual_tradeoff}
\end{figure}
Hereby, we additionally normalize by the number of residuals of the same type $N_k$ e.g., there exist $N_k=1$ gyroscope readings per keyframe, but in general $N_k>1$ reprojection residuals per keyframe.
\par
\subsection{Evaluation of Trade-off and Optimization}\label{sec_eval_opti}
A trade-off between gyroscope and camera-based residuals is adjusted by  $\beta_k = \gamma$ for residual $k$ belonging to gyroscope measurements and $\beta_k = (1-\gamma)$ for reprojection residuals, with $\gamma \in [0,1]$. 
We conduct experimental runs with optimization being triggered every 10 keyframes and a window size of 10.
Fig. \ref{fig_residual_tradeoff_subfigure_1} and \ref{fig_residual_tradeoff_subfigure_2} show the deviation of the SLAM algorithm from the ground-truth IR tracked poses for different values of $\gamma$.
Since we want to jointly optimize over all residuals, the cost emerging from different sensors should simultaneously decrease. The cost function difference $\Delta f$ is
\begin{equation}
\Delta f = f(\indizes{C}{T}\mathbf{T}_{n:m}^{(\text{after})}) - f(\indizes{C}{T}\mathbf{T}_{n:m}^{(\text{before})}),
\end{equation}
with evaluations before and after optimization.
Negative values indicate that this $k$-th part of the cost function was minimized.
In Fig. \ref{fig_residual_tradeoff_subfigure_3} we show the average cost difference for different residual types in one optimization step for different values of $\gamma$.
It exhibits a simultaneous cost decrease for values close to $\gamma = 0,4$.

On the other hand, we observe by inspecting the error trajectories in Fig. \ref{fig_residual_tradeoff_subfigure_1} and \ref{fig_residual_tradeoff_subfigure_2} that for values close to $0.5$ the optimization successfully incorporates the advantages of both sensors: The rotational error is low at the beginning, which is typical for camera-based tracking, but also stays down, which is possible due to utilizing the information of IMU sensors. 
The distribution of weights shows great potential in influencing the overall outcome of the algorithm. An optimal adjustment however, is a dynamic weight distribution, depending on the current conditions that influence the reliability of the sensor data.

\section{DISCUSSION}
Movement constraints and additional sensor information are necessary to provide stable pose estimates in dynamic environments. Absolute references are provided by the earth's gravity and magnetic field, whereas gyroscope and Lucas-Kanade tracks provide the odometry. For a more robust setup, however, descriptor matching of rigid features can improve the results of long-term SLAM.
\section{CONCLUSIONS}
This work shows that visual-inertial algorithms can be applied to dynamically challenging fields such as minimally invasive surgery.
This sets a new paradigm in the applicability of SLAM.
The gain of reliability is due to the fact that we use inherent motion constraints. Our results show for the use-case of minimally invasive surgery, that self localization is indeed feasible and hope that researchers can profit from our publicly available dataset.

%
%
%
%





%
%
\newpage
\bibliographystyle{IEEEtran}
\bibliography{../../Bib_Sources/mybib}

\end{document}